\documentclass{article}

\usepackage{arxiv}

\usepackage[utf8]{inputenc} % allow utf-8 input
\usepackage[T1]{fontenc}    % use 8-bit T1 fonts
\usepackage{hyperref}       % hyperlinks
\usepackage{url}            % simple URL typesetting
\usepackage{booktabs}       % professional-quality tables
\usepackage{amsfonts}       % blackboard math symbols
\usepackage{nicefrac}       % compact symbols for 1/2, etc.
\usepackage{microtype}      % microtypography
\usepackage{lipsum}
\usepackage{graphicx}
\graphicspath{ {./images/} }

\title{Exploratory Search with Sentence Embeddings}

\author{
 Austin Silveria \\
  California Polytechnic State University, San Luis Obispo\\
  San Luis Obispo, CA 93407 \\
  \texttt{silveria@calpoly.edu}
}

\begin{document}
\maketitle

\begin{abstract}
Exploratory search aims to guide users through a corpus rather than pinpointing exact information. We propose an exploratory search system based on hierarchical clusters and document summaries using sentence embeddings. With sentence embeddings, we represent documents as the mean of their embedded sentences, extract summaries containing sentences close to this document representation and extract keyphrases close to the document representation. To evaluate our search system, we scrape our personal search history over the past year and report our experience with the system. We then discuss motivating use cases of an exploratory search system of this nature and conclude with possible directions of future work.
\end{abstract}

\section{Introduction}
Traditionally, large corpora of documents are indexed and specific queries are used to locate information--documents matching these queries are returned and users can pinpoint the information they need. While this works well for extremely large corpora (the internet), this model does not support exploration because users need to know what they are looking for. Exploratory search aims to guide users through a large corpus of information by beginning with broad search goals, then assisting users in discovering information that is valuable to them. Table \ref{tab:search} differentiates primary search activities based on user goals as explored in a 2006 paper from the Communications of the ACM Journal \cite{10.1145/1121949.1121979}.

In this paper, we aim to support exploratory search activities by embedding a corpus's documents with pretrained sentence embeddings. Sentence embeddings are pretrained with general language tasks such as the Skip-Thought task \cite{kiros2015skipthought} in which an encoder and decoder are trained to reconstruct neighboring sentences of the given sentence--as a result, the encoder learns valuable information about general language that can be applied to other language tasks. We present a method that uses these embeddings to solve many problems related to exploratory search such as document organization, keyphrase extraction, summarization, and fuzzy-match searching and contribute a simple interface to explore the organized corpus.

Two major goals of our search interface are to support exploration of the corpus at varying levels of depth and to provide users with maximum transparency. Users should be able explore topics at a low level by being presented with documents that are very similar and at a high level by loosening the similarity  constraint. Users should also be able understand why certain documents are being presented together and visualize "where they are" in the embedding space. To satisfy these goals, we opted for an interface built to navigate a set of hierarchical clusters, as seen in Figure 1. The user exists at a position within the hierarchy at any given time and is presented with a number of options to further explore the corpus, such as opening a presented document, getting a summary of a presented document, moving in a particular direction, or performing a custom search.

\begin{table*}
  \caption{Search Activities \cite{10.1145/1121949.1121979}}
  \label{tab:search}
  \centering
  \begin{tabular}{ccl}
    \toprule
    Non-exploratory&Exploratory&Exploratory\\
    \midrule
    Lookup&Learn&Investigate\\
    \bottomrule
    Fact Retrieval&Knowledge acquisition&Accretion\\
    Known item search&Comprehension/Interpretation&Analysis\\
    Navigation&Comparison&Exclusion/Navigation\\
    Transaction&Aggregation/Integration&Synthesis\\
    Verification&Socialize&Evaluation\\
    Question answering&&Discovery\\
    &&Planning/Forecasting\\
    &&Transformation\\
\end{tabular}
\end{table*}

An overview of our methodology to support this interface is as follows: documents are represented by the mean of their embedded sentences. To extract keyphrases, we follow the strategy of EmbedRank \cite{bennanismires2018simple} and extract all noun phrases of each document, get their embedded representations, and score them based on their similarity to the document representation. Extracting a summary follows a similar process: embed each sentence of the document and score them based on their similarity to the document representation. Agglomerative clustering is applied to the document-level embedding space to find hierarchical groupings of documents. The clustered documents and simple summaries containing important sentences and keywords are precomputed to be quickly retrieved as the user navigates the document hierarchy.

\section{Backgound}

\subsection{Exploratory Search}
Exploratory search is a sub domain of information retrieval in which users are unsure of the exact information they want to retrieve. To aid in the development of exploratory search systems, a number of papers have sought to define what differentiates exploratory and non-exploratory search \cite{10.1145/1121949.1121979}\cite{doi:10.1002/asi.23617}. From Table \ref{tab:search}, it is clear that many different types of searching exist, each with their own unique needs. The possibility of an adaptive information retrieval system based on a search type classifier has been discussed in related literature \cite{doi:10.1002/asi.23617}. The authors gather metrics such as scroll length, query length, and number of clicks from users performing different types of search and report a search activity classification accuracy of 85\% across a subset of search activities that span "Lookup", "Learn", and "Investigate."

\begin{figure}[h]
  \label{fig:hierarchy}
  \centering
  \includegraphics[width=\linewidth]{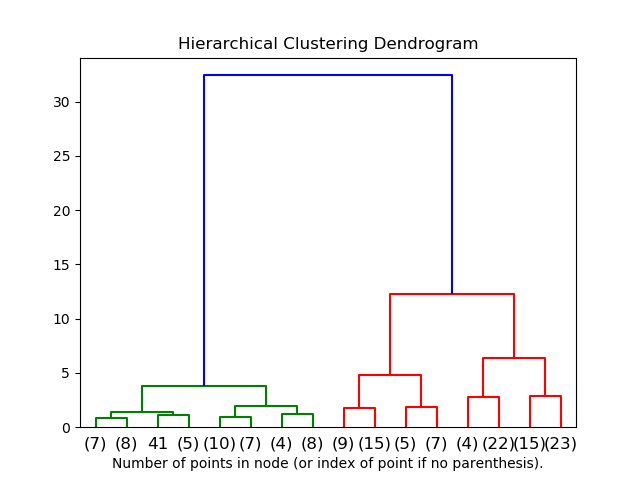}
  \caption{Hierarchical Clustering Visualization. Image from Scikit-Learn. (\url{https://bit.ly/379ou2c}).}
\end{figure}

Recommender systems are a type of exploratory search that have dominated information retrieval applications in recent years. YouTube released a paper in 2016 discussing their neural network architecture for recommending videos \cite{45530}--hundreds of candidates are chosen from YouTube's millions of videos based on a small number of features and those candidates are then ranked based on a larger number of features before being served to users in the order of ranking. The model aims to maximize watch time and has been extremely effective in delivering relevant video recommendations for users. Deep learning has dramatically improved the effectiveness of these recommender systems, especially at YouTube's scale, but has led to issues of transparency. In pure recommender systems that only produce ranked lists, it is often difficult to discern similarities and differences between items in the list, isolate subtopics within more general topics, and control the direction of exploration. Our purely hierarchical approach sacrifices the relevancy gained from deep learning in favor of transparency, but the best exploratory search systems will balance both.

\subsection{Sentence Embeddings}
\begin{figure*}[h]
  \centering
  \includegraphics[width=\linewidth]{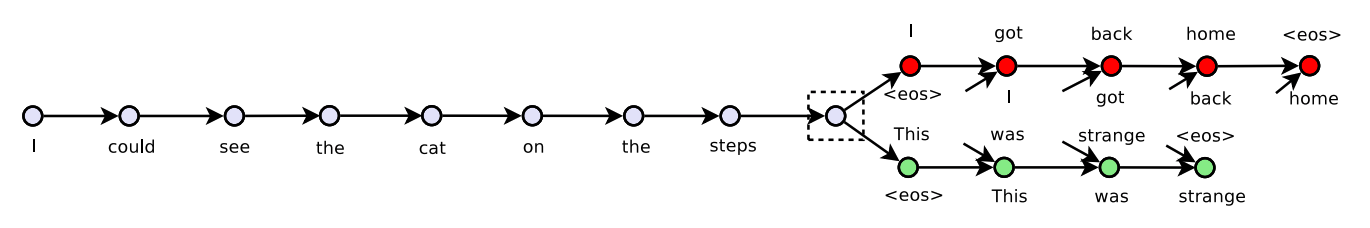}
  \caption{Skip-thoughts model \cite{kiros2015skipthought} reconstructing previous and next sentences}
  \label{skipthought}
\end{figure*}

Word2Vec \cite{mikolov2013efficient} introduced an efficient method to capture the meaning of words within vectors by using the current word to predict surrounding words. Previous work \cite{10.5555/944919.944966} used a three layer neural network with dense connections to estimate probabilities of words in context of other words as a language model. The intuition was correct in that training a model to predict a word's context results in the model capturing general language information, but the dense matrix multiplications were too computationally expensive. To solve this problem, Word2Vec removed the dense connections of the model, resulting in less representational power but the ability to scale to much larger datasets. Building on this efficient model, Skip-Thought Vectors \cite{kiros2015skipthought} extended the idea of predicting surrounding words to the sentence level. The task employed by this method is to reconstruct the previous and next sentences of the given sentence, as seen in Figure \ref{skipthought}. In our exploratory search system, we use Google's Universal Sentence Encoder \cite{cer2018universal}, a pretrained and freely available encoder for embedding sentences.

\subsection{Text Summarization}
The task of text summarization is primarily extractive or abstractive. In extractive summarization, the problem can be modeled as binary sentence classification--a sentence should either be included in the summary or not. This results in summaries that read like highlights of the full document, containing the most important information. Abstractive summarization uses a generative language model to form new text that summarizes the document at hand.

Many state of the art text summarization methods rely on supervised learning by training the model on human annotated summaries. BERTSum \cite{liu2019text} is a supervised text summarization method that fine tunes a pretrained language model for both extractive and abstractive summarization. By fine-tuning a pretrained language model, BERTSum can take advantage of the general information learned by the model in its pretraining phase of generating language. Unsupervised methods on the other hand, do not have access to human annotated summaries for training and generally rely on techniques to extract sentences closest to the document representation as a whole. The PacSum \cite{zheng2019sentence} model frames the sentences of the document as a graph. Directed edges are created between all sentences with weights as the similarities between the sentences pairwise. Graph ranking techniques such as PageRank \cite{ilprints422} can then be applied to this similarity matrix to find sentences that are most similar to all other sentences in the document. PacSum extends this sentence centrality idea by adjusting weights to sentence edges depending on if the edge points to a previous or next sentence in the document. Based on the inuition that sentences appearing earlier in the document should be more relevant to a summary, edges pointing to previous receive more weight.

\subsection{Keyphrase Extraction}
Keyphrase extraction is commonly a two stage process: parse candidate phrases from the document then rank them. A simple and effective method we use in this paper is EmbedRank \cite{bennanismires2018simple}. Candidate phrases contain zero or more adjectives followed by one or more nouns. The candidates are then embedded with sentence embeddings and are ranked by their distance to the document representation. The balance between keyphrase relevance and diversity is controlled by adjusting the ranking of a keyphrase based on how similar it is to already chosen keyphrases--ultimately, keyphrases are ranked highly if they are close to the document representation and far from already chosen keyphrases.

\section{Exploratory Search System}

The goal of our search system is to facilitate exploration of a corpus through comparison, aggregation, discovery, and exclusion of documents. By aggregating documents into groups that make intuitive sense to users, they can compare documents, exclude documents from further search, and discover new similarities and differences between documents. As a prerequisite of forming intuitive document groupings, it is necessary to have high quality representations of documents. Sentence embeddings offer a simple method of efficiently computing high quality document representations by averaging all embedded sentences of a document. Clustering algorithms can be applied to these representations to form document groupings with the specific algorithm being chosen to support the use case at hand.

Our exploratory search system begins with precomputation and storage of several pieces of data to support a real time interface during user interaction. Any use of sentence embeddings refers to Google's Universal Sentence Encoder \cite{cer2018universal}. Embedded representations of documents are first created by computing the mean of all embedded sentences per document and stored to disk. We then apply Scikit-Learn's agglomerative clustering to the document-level embedding space to hierarchically group documents and store the results on disk for easy access on application startup. Next, we extract document summaries using the following steps: embed all sentences, rank in ascending order by distance to the document representation, and take the closest three sentences as the summary. We attempted the PacSum \cite{zheng2019sentence} sentence centrality, but did not observe any improvement over our method, as discussed further in the evaluation section. Relevant keyphrases are extracted based on the method of EmbedRank \cite{bennanismires2018simple} as previously discussed: parse candidate phrases that contain zero or more adjectives followed by one or more nouns, embed all candidate phrases, and rank by distance to the document representation and distance to keyphrases already chosen. Summaries and keyphrases are also stored on disk and loaded into memory on application startup.

The key unit of our search interface is a cluster of documents. Each cluster contains any number of documents and a cluster is presented to users as a set of keyphrases extracted from the cluster’s documents and a sample of document titles. Upon interface startup, the user is initialized at the topmost position of the cluster hierarchy seen in Figure 1. At any position, users are presented the two children clusters of their current position, represented as left and right, and are given the following methods to interact with the hierarchy: refresh sampled keyphrases and document titles, move up, move left, move right, summarize a document, open a document, or type a search query. Document summaries consist of the extracted sentences closest to the document representation and extracted keyphrases. Upon typing a search query, the query is embedded with the Universal Sentence Encoder and a nearest neighbor search is performed to find the closest document. Users are then transported to a low level cluster containing this document and can explore the corpus from that position. Examples are shown in the evaluation section where we use our search system to explore a practical corpus.

\section{Experimental Design}

To evaluate our exploratory search system, we consider the webpages of our personal search history as a corpus. Using this corpus allows us to easily judge the formed document groupings and qualitatively discuss our experience of exploring topics interesting to us. In obtaining a corpus of webpage text for our search history, we first downloaded our Google data from Google Takeout. As a first preprocessing step, we grouped our website visits by domain to remove domains that did not have article form text such as Google Search, YouTube, GitHub, etc. We scraped the remaining website visit URLs to extract text by simply concatenating the <p> tags of each page. To remove duplicates, we applied a hash function to all pages’ extracted text and filtered duplicates. After filtering, we had 3900 website visits with title, URL, and text content that could be explored with our system.

We expect the exploratory search system to allow us to explore our topics of interest at varying levels of depth and help us understand our range of information consumption. We should easily be able to understand why sets of documents exist in the same cluster based on our prior knowledge of the language in the documents. The system should deliver enough transparency to enable us to clearly identify trends in the hierarchy as a whole and understand what they represent.

To evaluate our summarization method of extracting sentences close to the mean of all document sentences, we use the CNN/Daily Mail dataset. This dataset consists of a large number of medium length news articles with human annotated summaries for testing and training where ROUGE metric is used to evaluate the overlap between generated and human annotated summaries. We did not find many unsupervised methods evaluated on this dataset so we included many results of supervised methods for comparison.

\begin{table*}
  \caption{Exploratory Search Interface. Topmost position of hierarchy}
  \label{tab:top}
  \begin{tabular}{llcc}
    \toprule
    Left | 1394 Documents | data, Suge Knight, neural network &Right| 2507 Documents | DynamoDB, update npm, rsync\\
    \midrule
    Money Heist (TV Series 2017– ) - IMDb&How To Set Up a Firewall with UFW on Ubuntu 18.04 | Linuxize\\
    Transformers — transformers 2.6.0 documentation&How much IPFS Costs? - discuss.ipfs.io\\
    Richard Feynman - Wikipedia&Get started building an agile workflow | Atlassian\\
    Ontology (information science) - Wikipedia&AWS CodeBuild – Fully Managed Build Service\\
    Facebook AI&AWS SDK for Java\\
    Federated learning - Wikipedia&CSS Styling Images\\
    It's what the world wants : memes&Interfaces | Unreal Engine Documentation
\end{tabular}
\end{table*}

\section{Evaluation}
Our top level clusters using our search history corpus are illustrated in Table 2. The top two clusters are a split between technical software reading on the right with non-technical reading on the left. Table 3 represents the next two clusters presented after navigating “left” from the topmost cluster. These clusters can be seen as another split between artificial intelligence and non-artificial intelligence topics. We found the meanings of document clusters very intuitive and the movement throughout the cluster very natural as we were essentially “narrowing” our search. Navigating through the hierarchical clusters helped us maintain a visualization of where we were in the embedding space of our web reading--not only did the document clusters provide insight, but the interface of navigating one step at a time gave us powerful control over the direction of search.

Applied to the CNN/Daily Mail summarization dataset, our results are benchmarked against known supervised and unsupervised methods in Table 4. As is expected the supervised methods outperform the unsupervised methods, commonly by a few ROUGE points. BERTSum is currently the highest performing supervised method and PacSum outperforms Lead-3 (simply using the first three sentences as a summary) in the unsupervised category. As seen, our use of the PacSum model with Universal Sentence Encoder embeddings did not outperform sentence ranking based on distance to the mean and performed much worse than PacSum with skip-thought vectors, which is surprising because the Universal Sentence Encoder is based on the skip-thought task for training.

While the ROUGE is a good metric for summarization, we call to attention the question of whether summarizing is the best method for previewing documents in our exploratory search system. Extracting sentences close to the mean seems to extract a paragraph of interesting “highlights” with sentences that use keywords of the document. Opting for sentences that give a highlight of the document in question may prove more intriguing to users rather than a plain summary.

\section{Discussion}

Our search system is an efficient method to explore various corpora at varying levels of depth while also gaining insight into the structure of the corpus as a whole. In this section, we discuss future work that could build on top of our system. A natural next step in extending this system is to build a web interface in place of our command line implementation. With a web interface, the search system would be accessible to many users for purposes of testing or real use cases and would support easy upload of new corpora.

Scaling is another area where our implementation can be improved. The nearest neighbor searching functionality currently ranks a user’s query by embedded distance to all documents, and in the case of our search history corpus with only ~3900 documents this did not raise issues. However, with an internet-scale corpus containing millions of documents, a better solution will be needed to quickly search. Locality-sensitive hashing is a method that can drastically reduce the neighbor search speed by hashing similar items into the same “bucket.” By only computing distance to neighbors in the same buckets, it is possible to gain a massive speedup at the cost of losing a small amount of precision. In practice, the clusters our system already uses pose a great candidate to be used as the “buckets”--they group similar items and their hierarchical nature would allow dynamic adjusting of the neighbor search space size. In scaling to larger corpora, it would also be advantageous to have a method of incrementally updating the clusters as new documents are added, rather than performing the entire computation from the beginning. A possible solution would be to add new documents to existing clusters until their size warrants a split into smaller clusters.

\begin{table*}
  \caption{Exploratory Search Interface. Left from topmost position of hierarchy}
  \label{tab:left}
  \begin{tabular}{llcc}
    \toprule
    Left | 340 Documents | deep learning, ontology, parser&Right| 1054 Documents | Neuromancer, game theory, speeds\\
    \midrule
    Federated learning - Wikipedia&Mean Reversion Definition\\
    Gunning fog index - Wikipedia&Floyd Mayweather Jr. - Wikipedia\\
    AI Safety - Towards Data Science&Best racing games 2020 | PC Gamer\\
    ONNX expansion speeds AI development&OpenConnect VPN client.\\
    Facebook AI&Features | Spotify for Developers\\
    How Transformers Work - Towards Data Science&What are all the Jeff Dean facts? - Quora\\
    Introducing PySyft TensorFlow&Tactical shooter - Wikipedia
\end{tabular}
\end{table*}

\begin{table*}[bp]
  \caption{CNN/Daily Mail Summarization Results}
  \label{tab:down}
  \begin{tabular}{l c c c c}
    \toprule
    Model&R-1&R-2&R-L\\
    \midrule
    Supervised - BERTSum&43.85&20.34&39.90\\
    Supervised - TransformerExt&40.90&18.02&37.17\\
    Unsupervised - PacSum (BERT)&40.70&17.80&36.90\\
    Unsupervised - Lead-3&40.42&17.62&36.67\\
    Unsupervised - PacSum (skip-though vectors)&38.60&16.10&34.90\\
    Unsupervised - PacSum (Universal Sentence Encoder) (ours)&28.15&10.45&27.77\\
    Unsupervised - Distance to Mean (Universal Sentence Encoder) (ours)&30.08&10.98&29.41\\
\end{tabular}
\end{table*}

Currently, our implementation uses the ward linkage criteria which minimizes the variance of the clusters being merged. Experimenting with different linkage criteria, or using a combination of multiple, may result in discovery of new relationships between documents. For example, ward linkage may perform well in clustering based document type: Wikipedia vs. software documentation, but another criterion may group two documents of different types referring to the same thing in the same cluster. It would be interesting to explore the same corpus with multiple linkage criteria back to back.

At the pinnacle of exploratory search systems, the system would be able to generate abstractive summarizations of topics as a whole. Each position in the hierarchy would reference all documents in their “region” to write a general summary of the cluster as a whole. To summarize this much information, the model would likely need a form of memory to manage information about its entire cluster. A possible candidate for this task of long range summarization is a Memory Network \cite{weston2014memory}. This class of neural networks effectively learns to “write” memories that it finds useful to reference at inference time. Thus, the model could learn to write memories of each document to be used in crafting a general abstract summary. Another problem with abstractive summarization in this context, however, is the fact that it must be completely unsupervised. Similar to our method of extracting full sentences close to the mean, MeanSum \cite{chu2018meansum} is an abstractive model that learns to generate summaries that map close to the document representation in embedding space.

Another use case of an exploratory search system such as ours is deployment alongside a recommender system. The recommender system would continue to recommend the user content, and the exploratory search system could be applied to the corpus of consumed content, similar to our application of the system to our personal search history data. As users will already be familiar with the corpus, the document groupings will make intuitive sense and will provide them with a method of further understanding their interests and content consumption habits.

\section{Conclusion}

We have proposed an exploratory search system that uses sentence embeddings to embed the corpus and facilitate navigation through that space. By computing hierarchical clusters and summaries of documents, we support a real time interface for users to explore the corpus at hand. We represent documents as the mean embedding of all of their sentences, extract summaries containing sentences closest to the mean and extract keyphrases closest to the mean. These derived pieces of data are used to support a real time search interface that enables users to navigate the embedding space of their corpus. Sentence embeddings proved to be an effective “multitool” in natural language processing and gave strong baseline results on a variety of tasks. Going forward, researchers can build upon these baseline methods to create a more robust search system or even explore the same corpus with different sentence embeddings for evaluation purposes.

\bibliographystyle{unsrt}
\bibliography{explore}

\end{document}